\documentclass[runningheads]{llncs}

\usepackage{eccv}
\usepackage{eccvabbrv}
\usepackage{graphicx}
\usepackage{booktabs}
\usepackage[accsupp]{axessibility}
\usepackage{hyperref}
\usepackage{orcidlink}
\usepackage{algorithm}
\usepackage{multirow}
\usepackage{siunitx}
\usepackage{makecell}
\usepackage{mathtools}
\usepackage{tabularx,array}
\usepackage{listings}

\newcommand{\method}{{\textit{Resampling Forcing}}\xspace}

\definecolor{cmcomment}{RGB}{45,140,120}
\definecolor{cmfunc}{RGB}{214,40,120}
\definecolor{cmnum}{RGB}{40,60,200}

\lstdefinestyle{pseudocode}{
  basicstyle=\ttfamily\small,
  morecomment=[l]{\#},
  commentstyle=\color{cmcomment},
  keywords={def,for,in,return,with},
  keywordstyle=\bfseries,
  emph={denoise,logit_normal,randn_like,sample_t,mse_loss,no_grad,range,append,schedule,ode_step},
  emphstyle=\color{cmfunc},
  showstringspaces=false,
  columns=fullflexible,
  keepspaces=true,
  aboveskip=4pt, belowskip=2pt,
  literate=
    {-}{{{\color{cmnum}-}}}1 {+}{{{\color{cmnum}+}}}1 {*}{{{\color{cmnum}*}}}1,
}

\floatstyle{plain}
\restylefloat{algorithm}

\begin{document}

\title{End-to-End Training for Autoregressive Video Diffusion via Self-Resampling}
\titlerunning{Resampling Forcing}

\author{
Yuwei Guo\inst{1} \and
Ceyuan Yang\inst{2}$^{\dagger}$ \and
Hao He\inst{1} \and
Yang Zhao\inst{2} \and
Meng Wei\inst{3} \and \\
Zhenheng Yang\inst{3} \and
Weilin Huang\inst{2} \and
Dahua Lin\inst{1}
}
\authorrunning{Y.~Guo et al.}

\institute{The Chinese University of Hong Kong, Hong Kong, China \and ByteDance Seed, China \and ByteDance, China}

\maketitle

\makeatletter
\def\@thefnmark{}\@footnotetext{$^{\dagger}$~Corresponding author.}
\makeatother

\begin{abstract}
Autoregressive video diffusion models hold promise for world simulation but are vulnerable to exposure bias arising from the train–test mismatch. While recent works address this via post-training, they typically rely on a bidirectional teacher model or discriminator. To achieve an end-to-end solution, we introduce \method, a teacher-free framework that enables training autoregressive video models from scratch and at scale. Central to our approach is a self-resampling scheme that simulates inference-time model errors on history frames during training. Conditioned on these degraded histories, a sparse causal mask enforces temporal causality while enabling parallel training with frame-level diffusion loss. To facilitate efficient long-horizon generation, we further introduce history routing, a parameter-free mechanism that dynamically retrieves the top-$k$ most relevant history frames for each query. Experiments demonstrate that our approach achieves performance comparable to distillation-based baselines while exhibiting superior temporal consistency on longer videos owing to native-length training. See our \href{https://guoyww.github.io/projects/resampling-forcing/}{\textcolor{magenta}{Project Page}} for more details.
\end{abstract}

\section{Introduction}

Recent advances in generative video models have demonstrated strong potential for world modeling by approximating physical dynamics and predicting future states conditioned on current observations~\cite{videoworldsimulators2024, genie3, kang2024far, he2025pre}. Realizing this vision necessitates an autoregressive video generation paradigm that predicts the next frame conditioned on past context, thereby mirroring the \textit{strict causal} nature of the physical world. Beyond world simulation, such a paradigm enables a diverse array of applications, spanning game simulation~\cite{valevski2024diffusion, alonso2024diffusion, yu2025gamefactory, zhang2025matrix}, interactive content creation~\cite{lin2025autoregressive, shin2025motionstream}, and temporal reasoning~\cite{wiedemer2025video}.

\begin{figure}[t]
\centering \includegraphics[width=\linewidth]{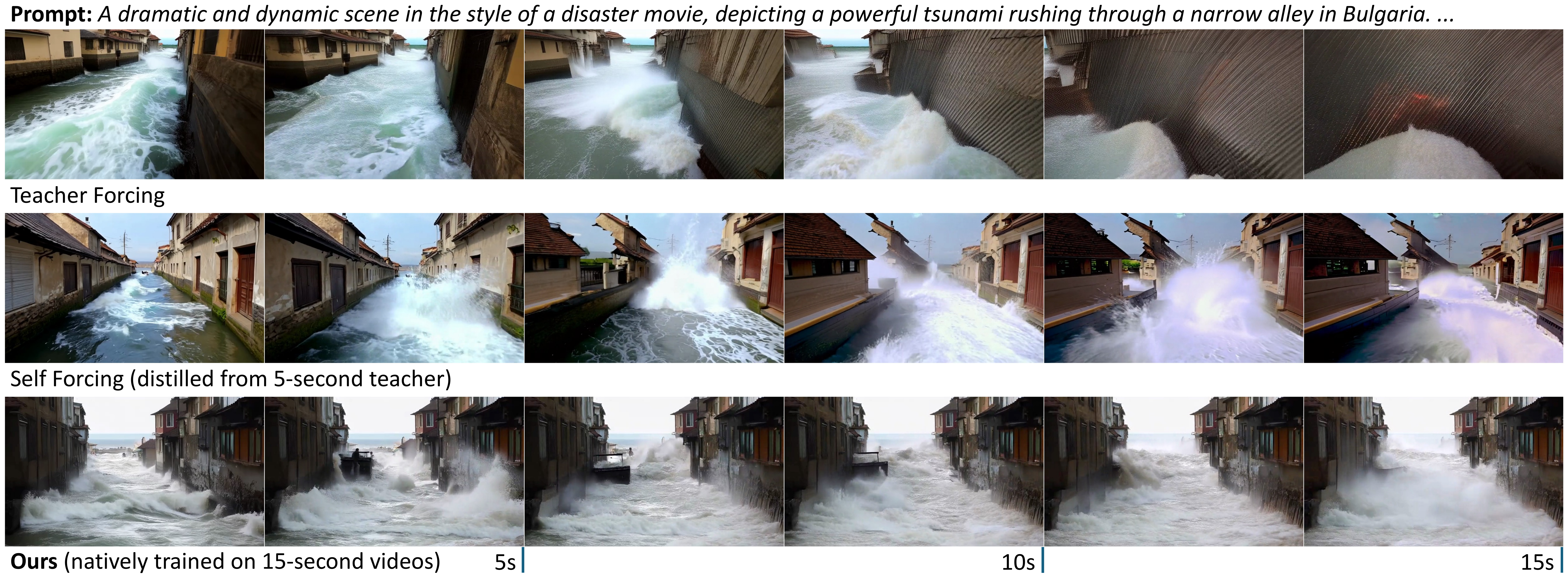}
\caption{We introduce \method, an end-to-end, teacher-free training framework for autoregressive video diffusion models. \textit{Top}: The teacher forcing accumulates errors and leads to video collapse. \textit{Middle}: Distilled from a short bidirectional teacher, Self Forcing suffers from the degraded quality on longer videos. \textit{Bottom}: Our method offers stable quality by native training on long videos.}
\label{fig:teaser}
\end{figure}

Despite its conceptual elegance, autoregressive video generation poses significant challenges. The primary hurdle is exposure bias~\cite{ning2023elucidating, schmidt2019generalization}: under teacher forcing, the model conditions on ground-truth histories during training, yet must rely on its own generated outputs during inference. This train–test mismatch can induce error accumulation, where small artifacts in model predictions are amplified across the autoregressive rollout, potentially leading to catastrophic video collapse~(see~\cref{fig:teaser} top). Furthermore, the ever-expanding historical context in autoregressive generation exacerbates attention complexity, posing practical obstacles for both training and inference over long horizons.

\begin{figure}[t]
\centering
\includegraphics[width=0.95\linewidth]{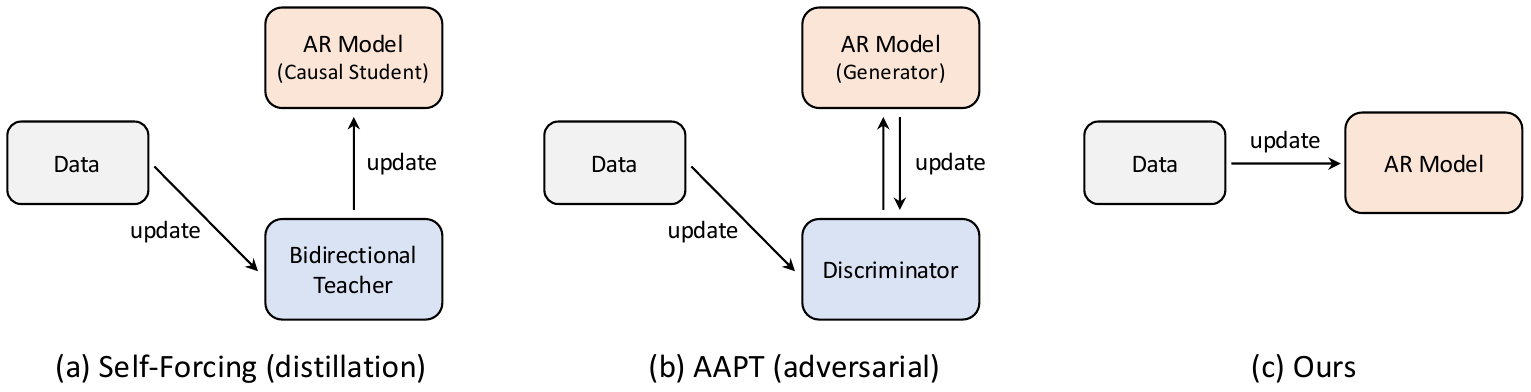}
\caption{
To obtain an autoregressive video diffusion model (AR model), (a)~Self Forcing~\cite{huang2025self} first trains a bidirectional teacher and then distills it into a causal student, while (b)~AAPT~\cite{lin2025autoregressive} adopts adversarial training by jointly optimizing the generator and discriminator.
(c)~We propose an end-to-end approach that directly trains the target model, without relying on any auxiliary model.
}
\label{fig:method_compare}
\end{figure}

To mitigate the train–test mismatch, recent works~\cite{huang2025self, lin2025autoregressive} employ post-training strategies aimed at aligning the generated video distribution with real data~(\cref{fig:method_compare}). For instance, Self Forcing~\cite{huang2025self} first autoregressively rolls out full videos, subsequently applying distillation or adversarial objectives to enforce distribution matching. By simulating inference during training, they reduce the discrepancy between training and test conditions. However, the reliance on a bidirectional teacher or an online discriminator impedes scalable training of autoregressive video models from scratch. A bidirectional teacher can also leak future information, compromising the strict temporal causality of the student model. Additionally, extensions to longer sequences typically use simple sliding-window attentions that disregard the varying importance of historical context, which may undermine long-term consistency.

In this work, we present \method, an end-to-end training framework for autoregressive video diffusion models. Drawing inspiration from the next-token prediction objective in LLMs, we condition each frame on its clean history and train in parallel via a per-frame diffusion loss under causal masking. We posit that, to mitigate error propagation and amplification, the model must be trained for robustness against input perturbations while retaining a clean prediction objective. To this end, the core of our method is a self-resampling mechanism: the model first induces errors in the history frames, then uses this degraded history to condition next-frame prediction. To simulate inference-time model errors, we autoregressively resample the latter segment of each frame’s denoising trajectory with the online model weights. This process is detached from gradient backpropagation to avoid shortcut learning. In addition, we introduce a history routing mechanism that dynamically retrieves the top-$k$ most relevant history frames via a parameter-free router, maintaining a near-constant attention complexity in long-horizon rollout.

Empirical results demonstrate that \method effectively mitigates error accumulation in autoregressive video diffusion models, achieving generation quality comparable to state-of-the-art distilled models. Leveraging native long-video training, our approach outperforms extrapolated baselines in longer video generation. Furthermore, our model exhibits stricter adherence to causal dependencies compared to distillation baselines. We also demonstrate that our history routing mechanism attains a sparse context with negligible quality loss, offering a viable memory design for long-horizon generation. We anticipate that this work will advance scalable training and long-term memory for future video world models.

\section{Related Works}

\noindent\textbf{Bidirectional Video Generation.} Bidirectional video generation refers to non-causal models that synthesize all frames jointly, allowing each frame to attend to both past and future context. Early efforts leveraged GANs~\cite{tulyakov2018mocogan, skorokhodov2022stylegan} or adapted UNet-based text-to-image diffusion models~\cite{blattmann2023align, guo2023animatediff, blattmann2023stable, bar2024lumiere}. Inspired by Sora~\cite{videoworldsimulators2024}, the field shifted toward 3D autoencoders and scalable Diffusion Transformers (DiT)~\cite{peebles2023scalable}, where all video tokens interact via self-attention, with text conditions injected through MMDiT-style fusion~\cite{esser2024scaling, kong2024hunyuanvideo, guo2025long} or separate cross-attention layers. State-of-the-art systems include commercial models such as Veo~\cite{veo}, Seedance~\cite{gao2025seedance}, Kling~\cite{kling}, as well as open-source ones like CogVideoX~\cite{yang2024cogvideox}, LTX-Video~\cite{hacohen2024ltx}, HunyuanVideo~\cite{kong2024hunyuanvideo}, and Wan~\cite{wan2025wan}. Our approach is built with a DiT-based video diffusion backbone.

\vspace{0.5em}
\noindent\textbf{Autoregressive Video Generation.} Recently, autoregressive video generation~\cite{li2024arlon, liu2024mardini, weng2024art, zhang2025generative, yuan2025lumos, ren2025autoregressive, henschel2025streamingt2v, gu2025starflow, liu2025infinitystar} has attracted increasing attention due to its potential in world and game simulation. It generates a video sequentially under a causal factorization, conditioning each frame on its historical context.

A pivotal challenge for autoregressive video diffusion is the train–test mismatch that leads to error accumulation~\cite{wang2025error}. Earlier attempts that directly use teacher forcing suffer from degraded quality as video length increases~\cite{gao2024ca2, hu2024acdit, zhang2025test}. To counteract this, prior work injects small noise into history frames to approximate inference-time degradation~\cite{valevski2024diffusion, weng2024art}. Another avenue adopts Diffusion Forcing~\cite{chen2025diffusion, song2025history, chen2025skyreels}, assigning each frame an independent noise level to enable conditioning at arbitrary noise during autoregressive rollout. Other works explore relaxing strict causality via a rolling denoising framework~\cite{ruhe2024rollingdiffusionmodels, teng2025magi, sun2025ar, xie2025progressive}. In this setup, video frames within a sliding window maintain non-decreasing noise levels and initiate generation when the preceding frame reaches a target timestep. Some works explore a plan-and-interpolate strategy that first generates a future keyframe and then interpolates the intermediate frames~\cite{zhang2025packing}.

Recently, Self Forcing-style post-training~\cite{huang2025self, lin2025autoregressive} has emerged as a promising solution for train–test alignment. It first autoregressively rolls out the entire video, then computes a holistic distribution matching loss. However, its reliance on online discriminators for adversarial loss~\cite{goodfellow2020generative} or pretrained bidirectional teachers for distillation~\cite{yin2024improved, yin2024one, zhou2024adversarial, zhou2024score} limits scalability and hinders training-from-scratch viability. While sharing the insight of inference simulation, our method uniquely supports end-to-end training without recourse to auxiliary models.

\vspace{0.5em}
\noindent\textbf{Conditioning on Model Predictions.} Conditioning the model on its own output is a strategy widely adopted to mitigate exposure bias in autoregressive systems. Scheduled Sampling~\cite{bengio2015scheduled, mihaylova2019scheduled, cen2024bridging} in language models exemplifies this approach, replacing portions of the ground-truth sequence with model-predicted tokens. In diffusion models, Self-Conditioning~\cite{chen2022analog} improves sample quality by conditioning the current denoising step on previous estimation. For video generation, Stable Video Infinity~\cite{li2025stable} adopts an error-recycling strategy to reduce drift in autoregressive long video inference.

\vspace{0.5em}
\noindent\textbf{Efficient Attention for Video Generation.} As high-dimensional spatiotemporal signals, videos require a large number of tokens to represent, making quadratic-cost attention a computational bottleneck. To alleviate complexity, numerous works explore efficient attention design for video generation. One line of work employs linear complexity attention~\cite{wang2025lingen, po2025long, chen2025sana, huang2025linvideo}. Others exploit the sparsity in the attention score, pruning less activated tokens via defined attention masks~\cite{xi2025sparse, xia2025training, sun2025vorta, zhang2025vsa, zhang2025spargeattn}. For instance, Radial Attention~\cite{li2025radial} observes a spatiotemporal decay and proposes a sparse mask that shrinks importance as temporal distance grows. Recent works also adapt advanced sparse-attention designs from LLMs~\cite{lu2025moba, yuan2025native}. For example, MoC~\cite{cai2025mixture} and VMoBA~\cite{wu2025vmoba} integrate Mixture of Block Attention~\cite{lu2025moba} into DiT blocks, where the key and value in attention are dynamically selected via a top-$k$ router. Our work integrates a similar routing mechanism, specifically tailored for handling long histories in autoregressive video generation.

\section{Method}

We begin by reviewing the background of autoregressive video diffusion models and analyzing the exposure bias issue in~\cref{sec:backgound}. We then present our \method algorithm for end-to-end training in~\cref{sec:resampling}, and the dynamic history routing for efficient long-horizon attention in~\cref{sec:history}.

\subsection{Background}
\label{sec:backgound}

\noindent\textbf{Definition.} Autoregressive~(AR) video diffusion models factorize video generation into inter-frame autoregression and intra-frame diffusion~\cite{ho2020denoising, sohl2015deep}. Specifically, given condition $c$, the joint distribution of an $N$-frame video sequence $\boldsymbol{x}^{1:N}$ is expressed as
\begin{equation}
p(\boldsymbol{x}^{1:N}|c) = \prod_{i=1}^N p(\boldsymbol{x}^i | \boldsymbol{x}^{<i}, c).
\end{equation}
To sample from each conditional distribution, the $i$-th frame $\boldsymbol{x}^i$ (denoted as $\boldsymbol{x}^i_0$ at timestep $t=0$) is synthesized by solving the reverse-time ODE starting from Gaussian noise $\boldsymbol{x}^i_1 \sim \mathcal{N}(\boldsymbol{0}, \boldsymbol{I})$ at timestep $t=1$ through
\begin{equation}
\label{eq:sampling}
\boldsymbol{x}^i = \boldsymbol{x}^i_1 + \int^0_1 \boldsymbol{v}_\theta(\boldsymbol{x}^i_t, \boldsymbol{x}^{<i}, t, c) \, \mathrm{d}t,
\end{equation}
which can be calculated via numerical solvers such as Euler. Here, the neural network $\boldsymbol{v}_\theta(\cdot)$ with parameter $\theta$ parameterizes the velocity field $\mathrm{d}\boldsymbol{x}^i_t / \mathrm{d}t$ and is conditioned on history frames $\boldsymbol{x}^{<i}$, \textit{i.e.}, $\mathrm{d}\boldsymbol{x}^i_t / \mathrm{d}t = \boldsymbol{v}_\theta(\boldsymbol{x}^i_t, \boldsymbol{x}^{<i}, t, c)$. Modern diffusion models typically employ the Diffusion Transformer~(DiT)~\cite{peebles2023scalable} architecture, where videos are patchified and processed via attention mechanisms~\cite{vaswani2017attention}. In practice, it is also common to generate a chunk of frames per autoregressive step. For simplicity, we refer to each chunk as a frame throughout this paper.

\vspace{0.5em}
\noindent\textbf{Teacher Forcing.} To train such sequence models, a common approach is teacher forcing, where the model is trained to predict the current frame $\boldsymbol{x}^i$ given its ground-truth history $\boldsymbol{x}^{<i}$. In Flow Matching~\cite{lipman2022flow}, the sample $\boldsymbol{x}^i_t$ at timestep $t$ is an interpolation between Gaussian noise $\boldsymbol{\epsilon}^i \sim \mathcal{N}(\boldsymbol{0}, \boldsymbol{I})$ and the clean frame $\boldsymbol{x}^i$ via
\begin{equation}
\label{eq:forward}
\boldsymbol{x}^i_t = (1 - t) \cdot \boldsymbol{x}^i + t \cdot \boldsymbol{\epsilon}^i.
\end{equation}
The network $\boldsymbol{v}_\theta(\cdot)$ is then trained to regress the velocity $\mathrm{d}\boldsymbol{x}^i_t / \mathrm{d}t = \boldsymbol{\epsilon}^i - \boldsymbol{x}^i$ by minimizing
\begin{equation}
\label{eq:objective}
\mathcal{L} = \mathbb{E}_{i, t, \boldsymbol{x}, \boldsymbol{\epsilon}} \left[ \lVert (\boldsymbol{\epsilon}^i - \boldsymbol{x}^i) - \boldsymbol{v}_\theta(\boldsymbol{x}^i_t, \boldsymbol{x}^{<i}, t, c) \rVert^2_2 \right].
\end{equation}
Here, $\boldsymbol{v}_\theta(\cdot)$ takes two separate sequences as inputs: the noisy frames $\boldsymbol{x}^i_t$ as diffusion samples and the noise-free ground-truth frames $\boldsymbol{x}^{<i}$. A causal mask restricts each frame to attend only to its clean history, enabling parallel training for all frames~(see~\cref{fig:method}~(b,c)). During inference, once a frame is generated, its clean features can be cached and reused for subsequent frame generation~(KV cache). Therefore, the number of attention queries remains constant, while the keys and values grow as the video becomes longer.

\begin{figure}[t]
\centering
\includegraphics[width=0.6\textwidth]{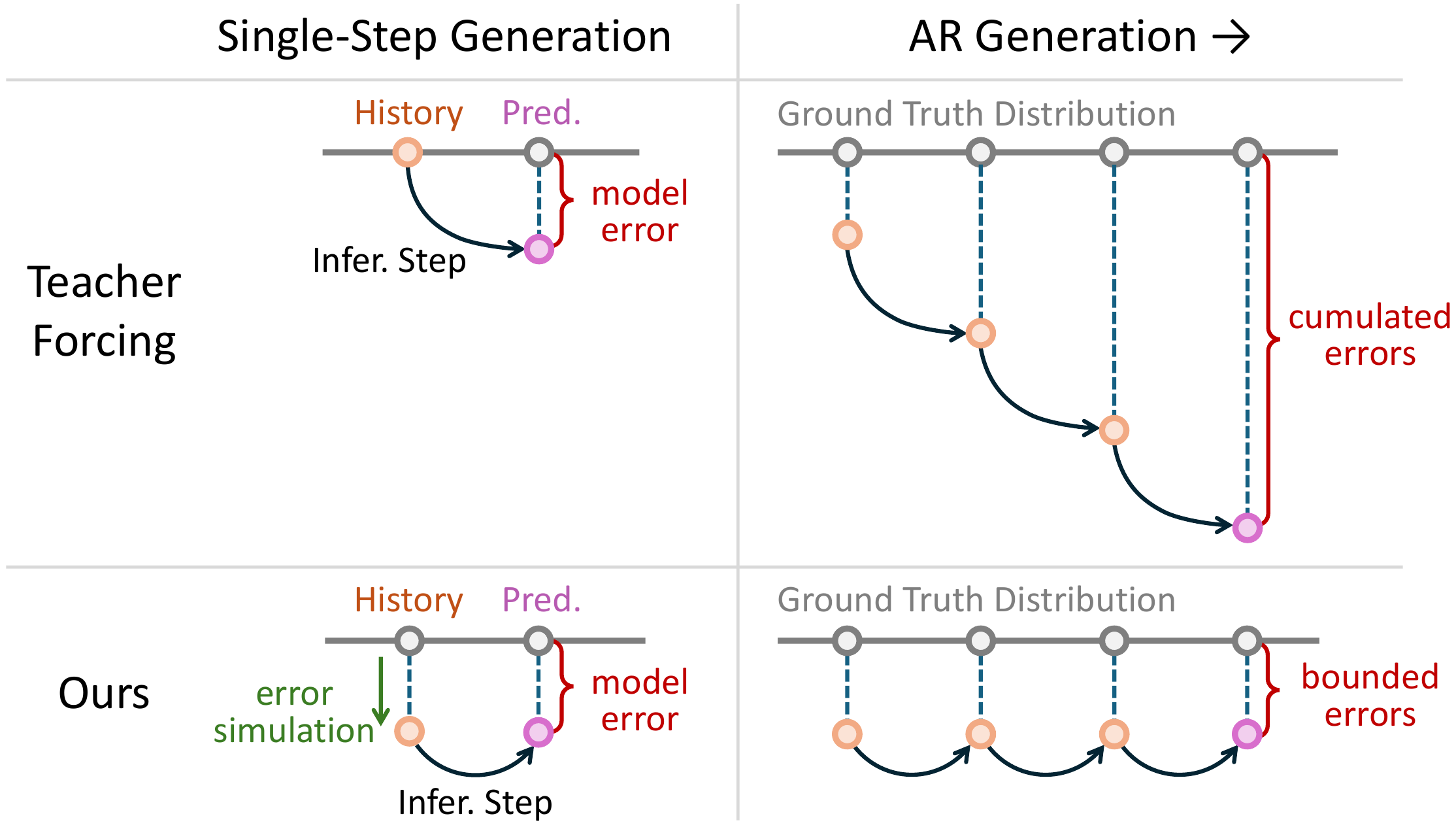}
\caption{\textbf{Error Accumulation.}
\textit{Top}: Models trained with ground-truth input add and compound errors autoregressively. \textit{Bottom}: We train the model on imperfect input with simulated model errors, stabilizing the long-horizon autoregressive generation. The gray circle represents the closest match in the ground-truth distribution. }
\label{fig:error}
\end{figure}

\vspace{0.5em}
\noindent\textbf{Error Accumulation.} Under teacher forcing, each frame's generation is conditioned on its ground-truth history. However, during inference, model predictions are inevitably \textit{imperfect}. In other words, model generations always have a non-zero discrepancy from the ground-truth distribution, which we refer to as \textit{model error}. The model trained on \textit{perfect} inputs will propagate and accumulate such errors through the autoregressive loop, leading to quality degradation and eventual failure in long-horizon rollouts~(see \cref{fig:error} top).

\subsection{Enhancing Error Robustness}
\label{sec:resampling}

As analyzed above, the failure of teacher forcing stems from the distributional mismatch between training and inference inputs, driven by irreducible model errors. With finite model capacity and training samples, eliminating such error is intractable. Instead, we propose training the model to condition on degraded histories while maintaining error-free targets for prediction. This relaxes the model from strict adherence to input conditions and enables correction of input errors. As illustrated at the bottom of~\cref{fig:error}, although the model predictions remain imperfect, errors no longer compound. Instead, the errors are stabilized to a near-constant level over the autoregressive process.

\begin{figure}[t]
\centering
\includegraphics[width=0.8\linewidth]{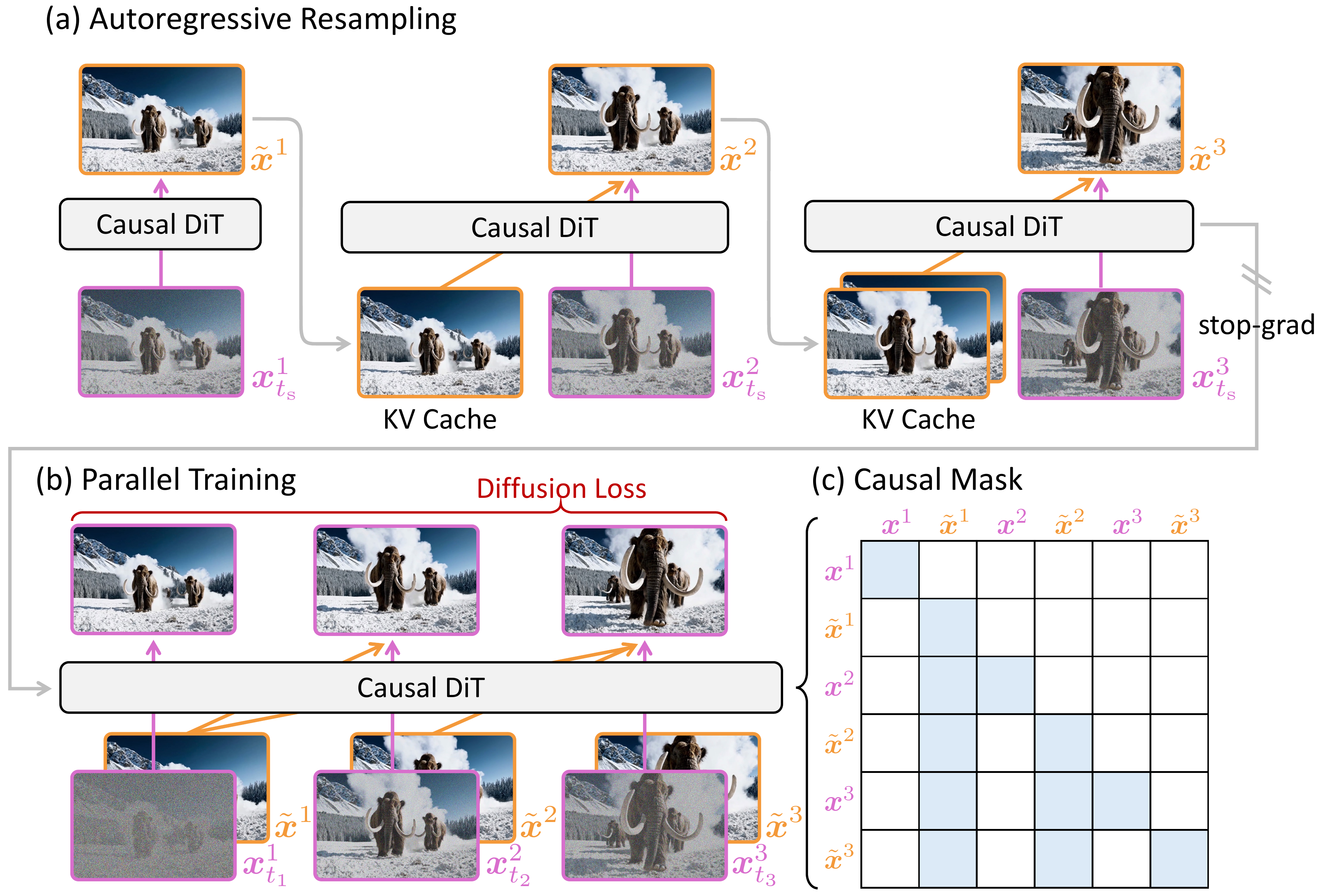}
\caption{\textbf{\method.}
(a)~To simulate inference-time model error, we add noise to clean videos up to a sampled timestep~$t_s$, then use the online model weights to autoregressively complete the remaining denoising steps. (b)~The model is parallel trained with frame-level diffusion loss. (c)~A sparse causal mask restricts each frame to attend only to its clean history frames. }
\label{fig:method}
\end{figure}

\vspace{0.5em}
\noindent\textbf{Simulating the Model Error.} To train error-robust autoregressive models, we must simulate inference-time model errors on the input conditions. We pursue an end-to-end, teacher-free approach to achieve this, prioritizing simplicity and scalability. There are two major factors that drive the distribution shift in autoregressive diffusion: (1)~intra-frame generation errors that arise from imperfect score estimation and discretization, which mainly affect high-frequency details~\cite{wang2025error, falck2025fourier}; and (2)~inter-frame accumulated errors that propagate through the autoregressive loop~\cite{wang2025error, yin2024slow}.

To simulate errors from both aspects, we introduce \textit{autoregressive self-re\-sampling} on the history condition. To mimic intra-frame errors, we resample the latter denoising trajectory, where high-frequency details are typically synthesized. Specifically, we corrupt ground-truth video frame $\boldsymbol{x}^i$ to a sampled timestep $t_s \in (0, 1)$ via~\cref{eq:forward} to obtain $\boldsymbol{x}^i_{t_s}$. Subsequently, we employ the online model $\boldsymbol{v}_\theta(\cdot)$ to complete the remaining denoising steps and produce a degraded noise-free frame $\tilde{\boldsymbol{x}}^i$ that contains model errors. The timestep $t_s$ controls how close $\tilde{\boldsymbol{x}}^i$ is to its ground-truth version $\boldsymbol{x}^i$. To mimic inter-frame error accumulation, we resample each frame autoregressively, conditioning on degraded history frames $\tilde{\boldsymbol{x}}^{<i}$~(see~\cref{fig:method}~(a)), \textit{i.e.},
\begin{equation}
\label{eq:resampling}
\tilde{\boldsymbol{x}}^i = \boldsymbol{x}^i_{t_s} + \int^0_{t_s} \boldsymbol{v}_\theta(\boldsymbol{x}^i_t, \tilde{\boldsymbol{x}}^{<i}, t, c) \, \mathrm{d}t.
\end{equation}
Using online model weights ensures that the error distribution evolves alongside training, thereby compelling the model to continuously learn to correct its current imperfections. Gradients are detached from this process to prevent shortcut learning. In practice, this procedure can be efficiently implemented with KV cache.

\vspace{0.5em}
\noindent\textbf{Sampling Simulation Timestep.} The timestep $t_s$ in~\cref{eq:resampling} governs the trade-off between history faithfulness and error correction flexibility. A small $t_s$ yields low resampling strength, resulting in a degraded sample $\tilde{\boldsymbol{x}}^i$ that closely resembles its ground-truth $\boldsymbol{x}^i$. This encourages the model to stay faithful to the history frames and risks error accumulation~(teacher forcing is a limiting case where $t_s=0$). On the other hand, a large $t_s$ grants greater flexibility for error correction but raises the chance of content drift, as the model is permitted to deviate significantly from the historical context. Consequently, the distribution of $t_s$ should concentrate density on intermediate values while suppressing extremes. To model this, we choose to sample $t_s$ from a logit-normal distribution $\mathrm{LogitNormal}(0, 1)$ that satisfies the above properties:
\begin{equation}
\text{logit}(t_s) \sim \mathcal{N}(0, 1).
\end{equation}
Generally, stronger models induce fewer errors, allowing for a greater emphasis on low resampling strength, and vice versa. To bias the distribution of $t_s$, inspired by~\cite{esser2024scaling}, we apply a timestep shifting with parameter $s$ after sampling $t_s$ from the standard logit normal distribution via
\begin{equation}
\label{eq:timestep_shift}
t_s \leftarrow \frac{s \cdot t_s}{1 + (s-1) \cdot t_s}.
\end{equation}
In implementation, we set $s < 1$ to put more weight on the low-noise region. After resampling, we use the degraded video $\tilde{\boldsymbol{x}}^{1:N}$ as history condition, and use the ground-truth video $\boldsymbol{x}^{1:N}$ as the training objective. The complete pseudo code for \method is shown in~\cref{alg:train}.

\vspace{0.5em}
\noindent\textbf{Teacher Forcing Warmup.} In the initial training phase, the model has not yet converged to the causal architecture and is incapable of generating meaningful content autoregressively. The model errors at this stage are dominated by random initialization rather than specific intra-frame imperfections or inter-frame accumulation. Therefore, performing history self-resampling can lead to uninformative learning signals and will hinder convergence. We therefore first warm up the model using teacher forcing. Once the model acquires basic autoregressive capabilities (though imperfect), we transition to \method and continue training.

\begin{algorithm}[t]
\centering
\begin{minipage}{\linewidth}
\refstepcounter{algorithm}\label{alg:train}%
\kern1pt\hrule height 0.9pt\kern3pt
\noindent\textbf{Algorithm \thealgorithm}\hspace{0.7em}\method\par
\kern2pt\hrule height 0.4pt\kern2pt
\begin{lstlisting}[style=pseudocode, aboveskip=2pt, belowskip=1pt]
# v(x_t, hist, t, c): causal diffusion model
# x: clean video frames; c: condition
# x_hat: resampled history
# s: shift parameter

def denoise(v, z, hist, t_s, c):
    for t, t_next in schedule(t_s, 0):
        z = ode_step(v, z, hist, t, t_next, c)
    return z

# (a) autoregressive resampling
t_s = logit_normal()
t_s = s * t_s / (1 + (s - 1) * t_s)    # timestep shift
with no_grad():
    x_hat = []
    for i in range(N):
        z = (1 - t_s) * x[i] + t_s * randn_like(x[i])
        x_hat.append(denoise(v, z, x_hat, t_s, c))

# (b) diffusion loss
t = sample_t(N)
e = randn_like(x)
z = (1 - t) * x + t * e
target = e - x

pred = v(z, x_hat, t, c)
loss = mse_loss(pred, target)
\end{lstlisting}
\kern2pt\hrule height 0.9pt
\end{minipage}
\end{algorithm}

\subsection{Routing History Context}
\label{sec:history}

\begin{figure}[t]
\centering
\includegraphics[width=0.6\textwidth]{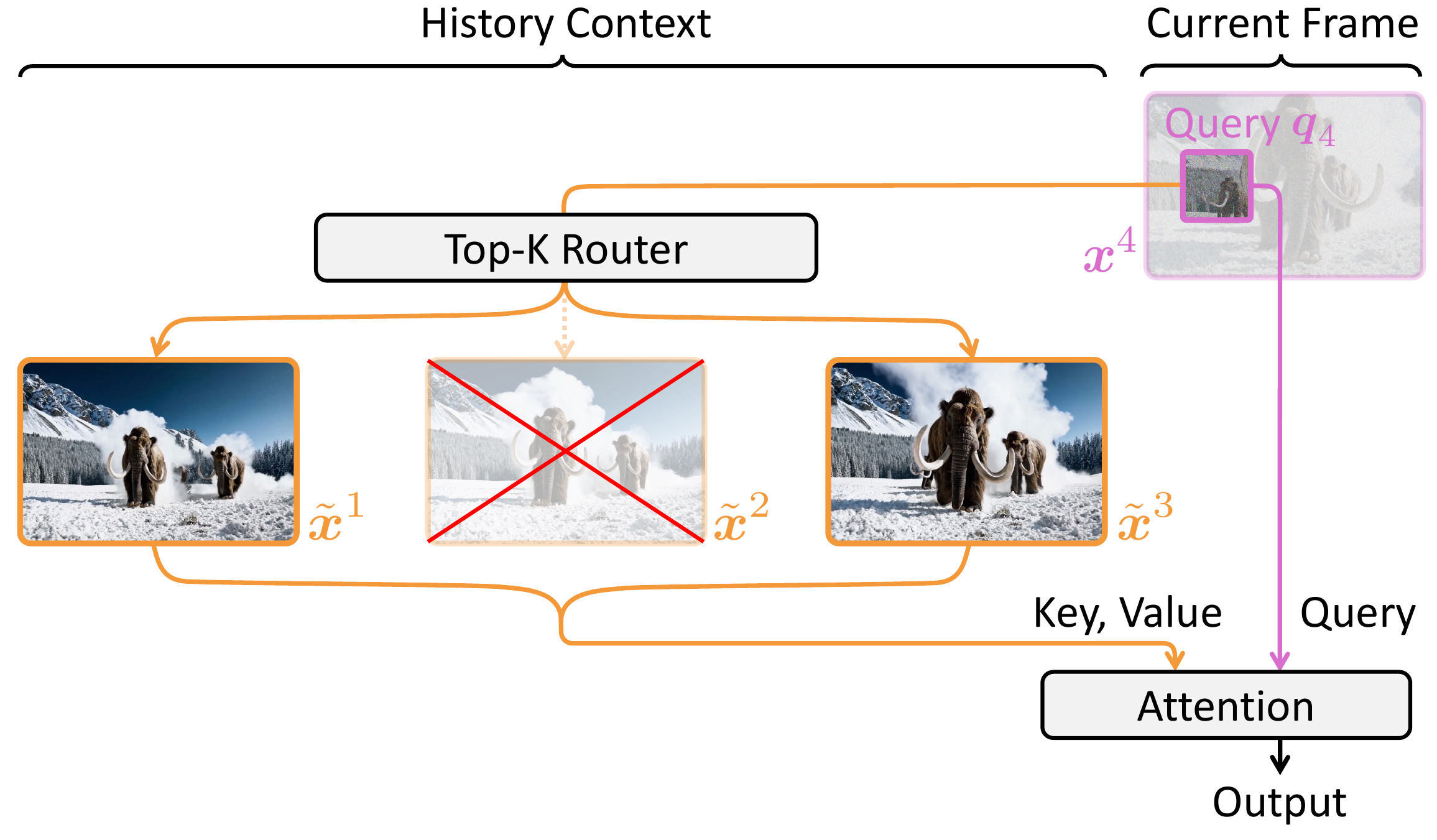}
\caption{\textbf{History Routing Mechanism.}
Our routing mechanism dynamically selects the top-$k$ important frames to attend. In this illustration, we show a $k=2$ example, where only the 1st and 3rd frames are selected for the 4th frame's query token $\boldsymbol{q}_4$.}
\label{fig:route_history}
\end{figure}

In autoregressive generation, the number of history frames grows as the video becomes longer. This decelerates generation for subsequent frames, as dense causal attention necessitates attending to the entire historical context. To resolve this, a common solution is to restrict the attention receptive field to a local sliding window~\cite{huang2025self, lin2025autoregressive, yang2025longlive}. However, this approach compromises long-term dependency, sacrificing global consistency and exacerbating the drifting problem.

To maintain stable attention complexity, we opt to \textit{optionally} replace the dense causal attention with a dynamic routing mechanism, inspired by the advanced sparse attention in LLMs~\cite{lu2025moba, yuan2025native}. Specifically, for the query token $\boldsymbol{q}_i$ of the $i$-th frame, rather than attending to the full history, we dynamically retrieve and attend to the top-$k$ most relevant history frames~(see~\cref{fig:route_history}), \textit{i.e.},
\begin{equation}
\mathrm{Attention}(\boldsymbol{q}_i, \boldsymbol{K}_{<i}, \boldsymbol{V}_{<i}) = \mathrm{Softmax} \left( \frac{\boldsymbol{q}_i \boldsymbol{K}^\top_{\Omega(\boldsymbol{q}_i)}}{\sqrt{d}} \right) \cdot \boldsymbol{V}_{\Omega(\boldsymbol{q}_i)},
\end{equation}
where $\Omega(\boldsymbol{q}_i)$ is set of selected indices of $k$ history frames for query $\boldsymbol{q}_i$. For selection metric, we use the dot product of $\boldsymbol{q}_i$ and a frame descriptor $\phi(\boldsymbol{K}_j)$~(for $j$-th frame), \textit{i.e.},
\begin{equation}
\Omega(\boldsymbol{q}_i) = \arg\max_{|\Omega^*| = k} \sum_{j\in\Omega^*} \left(\boldsymbol{q}^\top_i \phi(\boldsymbol{K}_j) \right).
\end{equation}
Following~\cite{lu2025moba, cai2025mixture, wu2025vmoba}, we use mean pool as the descriptor transformation $\phi(\cdot)$ since it adheres to the attention score computation and is parameter-free. This reduces the per-token attention complexity from linear $\mathcal{O}(L)$ to constant $\mathcal{O}(k)$ as the number of history frames $L$ grows, achieving an attention sparsity of $1 - k/L$.

Notably, while $k$ may be set small for high sparsity, the routing mechanism operates in a \textit{head-wise} and \textit{token-wise} manner, implying that tokens across different attention heads and spatial locations can route to distinct history mixtures, and collectively yield a receptive field significantly larger than $k$ frames.

In implementation, following MoBA~\cite{lu2025moba}, we adopt an efficient two-branch attention that fuses an intra-frame pathway and a sparse history pathway via a global log-sum-exp. In the intra-frame branch, each query attends only to tokens within its own frame. In the history branch, we select up to top-$k$ relevant past frames for each query token. Both branches are implemented efficiently with the \texttt{flash\_attn\_varlen\_func()} interface from FlashAttention~\cite{dao2022flashattention, dao2023flashattention2}. Outputs are combined by aligning their log-sum-exp terms, yielding a result equivalent to a single softmax over the union of keys.

\section{Experiments}

\noindent\textbf{Model.} We build our method upon \texttt{Wan2.1-1.3B}~\cite{wan2025wan} architecture.
We modify timestep conditioning to support per-frame noise levels, and implement the sparse causal attention in~\cref{fig:method}~(c) with \texttt{flex\_attention()} in PyTorch, incurring no additional parameters. Following~\cite{huang2025self,yang2025longlive,cui2025self}, we use a chunk size of 3 latent frames as the autoregressive unit.

\vspace{0.5em}
\noindent\textbf{Training.} After switching to causal attention, the model was trained on $5\,\mathrm{s}$ videos with the teacher forcing objective for 10K steps to warm up, then transitioned to \method and trained sequentially on $5\,\mathrm{s}$ and $15\,\mathrm{s}$~(249 frames) videos, for 15K and 5K steps respectively. We then fine-tuned with sparse history routing enabled for 1.5K iterations on $15\,\mathrm{s}$ videos. The training batch size is $64$ and the learning rate for the AdamW optimizer is $5\times10^{-5}$. We set the timestep shifting factor $s=0.6$ (\cref{sec:resampling}), and $k=5$ in top-$k$ history routing~(\cref{sec:history}). For efficiency, we use a 1-step Euler solver for history resampling (\cref{eq:resampling}).

\vspace{0.5em}
\noindent\textbf{Inference.} We use the same inference settings for all generated frames. We use an Euler sampler with 32 steps and a timestep shifting factor of $5.0$. The classifier-free guidance scale is $5.0$ for all frames.

\subsection{Comparisons}

\noindent\textbf{Baselines.} We compare our method with recent autoregressive video generation baselines, including SkyReels-V2~\cite{chen2025skyreels}, MAGI-1~\cite{teng2025magi}, NOVA~\cite{deng2024autoregressive}, Pyramid Flow~\cite{jin2024pyramidal}, CausVid~\cite{yin2024slow}, Self Forcing~\cite{huang2025self}, and a concurrent work, LongLive~\cite{yang2025longlive}. Notably, SkyReel-V2 operates as a clip-level autoregressive model, generating $5\,\mathrm{s}$ video segments sequentially. MAGI-1 relaxes the strict causal constraint, initiating next-chunk denoising prior to the completion of the current chunk's generation. LongLive rolls out longer videos and takes $5\,\mathrm{s}$ sub-clips and computes distillation loss with teacher models.

\begin{figure}[t]
\centering \includegraphics[width=\linewidth]{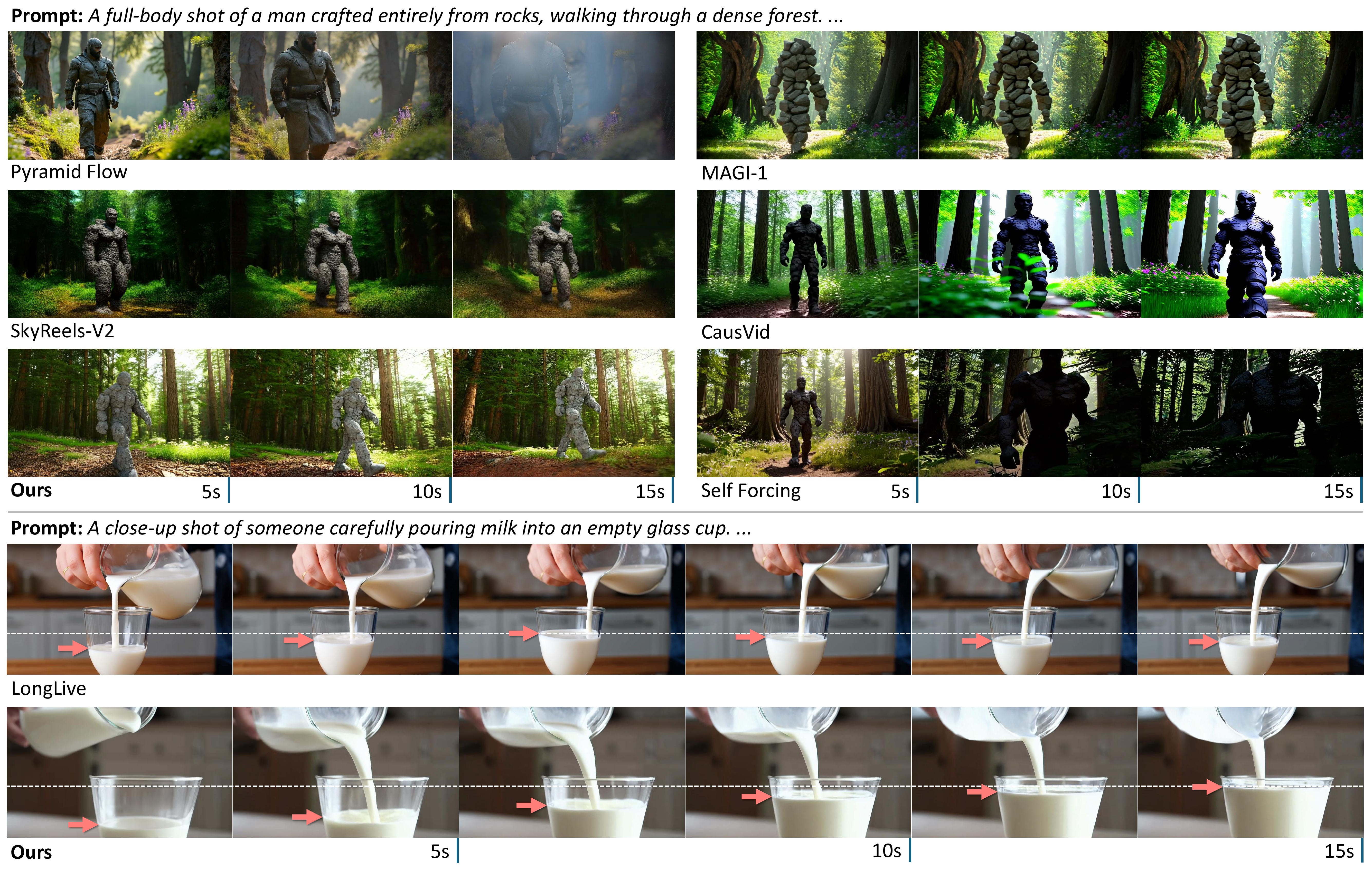}
\caption{\textbf{Qualitative Comparisons.}
\textit{Top}: We compare with representative autoregressive video generation models, showing our method's stable quality on long video generation. \textit{Bottom}: Compared with LongLive~\cite{yang2025longlive} that distilled from a short bidirectional teacher, our method exhibits better causality. We use dashed lines to denote the highest liquid level, and red arrows to highlight the liquid level in each frame. }
\label{fig:qual_compare}
\end{figure}

\vspace{0.5em}
\noindent\textbf{Qualitative Comparison.} We provide a visual qualitative comparison across different methods in~\cref{fig:qual_compare} on generated 15-second videos.

In the upper panel, we compare the visual quality, observing that most strict autoregressive models~(\textit{e.g.}, Pyramid Flow~\cite{jin2024pyramidal}, CausVid~\cite{yin2024slow}, and Self Forcing~\cite{huang2025self}) exhibit error accumulation, manifested as progressive degradation in color, texture, and overall sharpness.
By contrast, approaches that relax strict causality (MAGI-1~\cite{teng2025magi}) or use large autoregressive chunks (SkyReels-V2~\cite{chen2025skyreels}) alleviate long-horizon degradation. However, these relaxed settings compromise intrinsic advantages of strict autoregression, such as per-frame interactivity and faithful causal dependency for future-state prediction. Within the strict autoregressive paradigm, our method demonstrates superior robustness in long-term visual quality compared to baselines.

In the lower panel, we further compare with LongLive~\cite{yang2025longlive}, which first generates long videos and then performs sub-clip distillation using a short-horizon teacher. Although LongLive attains strong long-range visual quality, we observe that distillation from a short bidirectional teacher fails to ensure strict causality, even with a temporally causal student architecture. In the ``milk pouring'' example in \cref{fig:qual_compare}, LongLive produces a liquid level that rises and then falls despite continuous pouring, which violates physical laws. By contrast, our model maintains strict temporal causality: the liquid level monotonically increases while the source container empties. We attribute this non-causal behavior to two factors. First, the bidirectional teacher is inherently non-causal, allowing future information to influence earlier frames via attention, thereby leaking future context to the student during distillation. Second, sub-clip distillation emphasizes local appearance quality and neglects global causality. Conversely, our training strictly precludes information leakage from the future.

\vspace{0.5em}
\noindent\textbf{Quantitative Comparison.} We evaluate methods using the automatic metrics provided by VBench~\cite{huang2024vbench}. All models are prompted to generate 15-second videos, which we partition into three segments and evaluate them separately to better assess long-term quality. Results are summarized in \cref{tab:quant_comparison}. As evidenced in the table, our method maintains comparable visual quality and superior temporal quality on all video lengths to baselines. On longer video lengths, our method's performance also matches the long video distillation baseline LongLive. Given that the distill-based methods~(\textit{i.e.}, CausVid~\cite{yin2024slow}, Self Forcing~\cite{huang2025self}, and LongLive~\cite{yang2025longlive}) necessitate a pretrained 14B-parameter bidirectional teacher, our method offers substantial efficiency and practicality for training autoregressive video models. Moreover, the history routing mechanism achieves an attention sparsity of $75\%$ while incurring only a negligible drop relative to the dense-attention baseline, demonstrating its strong potential for long-horizon generation under constrained compute and memory budgets.

\begin{table*}[t]
\centering
\footnotesize
\renewcommand{\arraystretch}{0.9}
\caption{\textbf{Quantitative Comparisons.}
We split the generated 15-second videos into three parts, \textit{i.e.}, $0\text{--}5\,\mathrm{s}$, $5\text{--}10\,\mathrm{s}$, and $10\text{--}15\,\mathrm{s}$, and separately evaluate the videos with VBench~\cite{huang2024vbench}.}
\label{tab:quant_comparison}
\resizebox{\linewidth}{!}{
\begin{tabular}{lcc ccc ccc ccc}
\toprule \multirow{2}{*}{Method} & \multirow{2}{*}{\#Param} & \multirow{2}{*}{Teacher} & \multicolumn{3}{c}{Length = $0\text{--}5\,\mathrm{s}$} & \multicolumn{3}{c}{Length = $5\text{--}10\,\mathrm{s}$} & \multicolumn{3}{c}{Length = $10\text{--}15\,\mathrm{s}$} \\
& & & \textit{Temp.} & \textit{Visual} & \textit{Text} & \textit{Temp.} & \textit{Visual} & \textit{Text} & \textit{Temp.} & \textit{Visual} & \textit{Text} \\
\midrule SkyReels-V2~\cite{chen2025skyreels} & 1.3B & - & 81.93 & 60.25 & 21.92 & 84.63 & 59.71 & 21.55 & 87.50 & 58.52 & 21.30 \\
MAGI-1~\cite{teng2025magi} & 4.5B & - & 87.09 & 59.79 & 26.18 & 89.10 & 59.33 & 25.40 & 86.66 & 59.03 & 25.11 \\
NOVA~\cite{deng2024autoregressive} & 0.6B & - & 87.58 & 44.42 & 25.47 & 88.40 & 35.65 & 20.15 & 84.94 & 30.23 & 18.22 \\
Pyramid Flow~\cite{jin2024pyramidal} & 2.0B & - & 81.90 & 62.99 & 27.16 & 84.45 & 61.27 & 25.65 & 84.27 & 57.87 & 25.53 \\
CausVid~\cite{yin2024slow} & 1.3B & \texttt{Wan-14B(5s)} & 89.35 & 65.80 & 23.95 & 89.59 & 65.29 & 22.90 & 87.14 & 64.90 & 22.81 \\
Self Forcing~\cite{huang2025self} & 1.3B & \texttt{Wan-14B(5s)} & 90.03 & 67.12 & 25.02 & 84.27 & 66.18 & 24.83 & 84.26 & 63.04 & 24.29 \\
LongLive~\cite{yang2025longlive} & 1.3B & \texttt{Wan-14B(5s)} & 81.84 & 66.56 & 24.41 & 81.72 & 67.05 & 23.99 & 84.57 & 67.17 & 24.44 \\
\midrule Ours (75\% sparsity) & 1.3B & - & 90.18 & 63.95 & 24.12 & 89.80 & 61.95 & 24.19 & 87.03 & 61.01 & 23.35 \\
Ours & 1.3B & - & 91.20 & 64.72 & 25.79 & 90.44 & 64.03 & 25.61 & 89.74 & 63.99 & 24.39 \\
\bottomrule
\end{tabular}
}
\end{table*}

\subsection{Analytical Studies}

\begin{figure}[t]
\centering
\includegraphics[width=\linewidth]{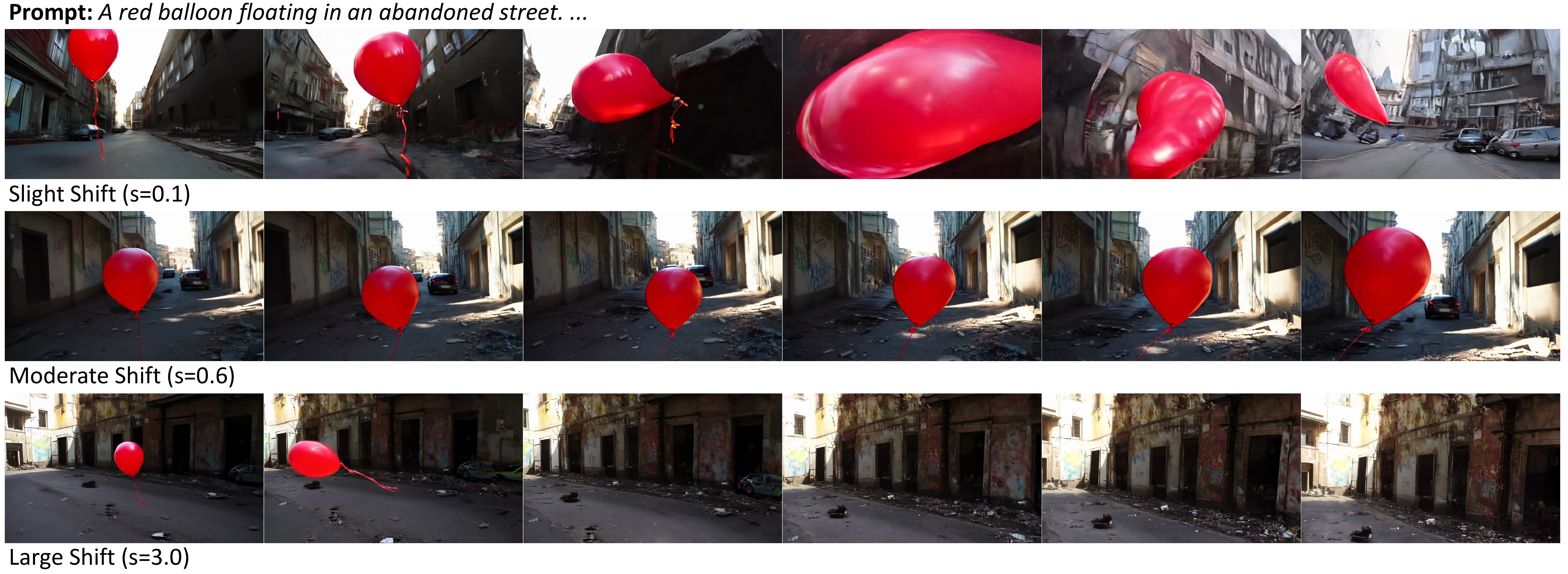}
\caption{\textbf{Comparing Timestep Shifting.} A moderate shifting scale for resampling timestep $t_s$ is necessary to balance between error accumulation and content drift.}
\label{fig:ablation_shift}
\end{figure}

\begin{table}[t]
\centering
\footnotesize
\renewcommand{\arraystretch}{0.9}
\setlength{\tabcolsep}{10pt}
\caption{\textbf{Error Simulation Strategies.} We compare different variants and find that autoregressive resampling yields the best quality.}
\label{tab:ablation}
\resizebox{0.65\linewidth}{!}{
\begin{tabular}{lccc}
\toprule
\multirow{2}{*}{Simulation Strategies} & \multicolumn{3}{c}{Length = $0\text{--}15\,\mathrm{s}$} \\
& \textit{Temp.} & \textit{Visual} & \textit{Text} \\
\midrule
Noise Augmentation & 87.15 & 61.90 & 21.44 \\
Parallel Resampling & 88.01 & 62.51 & 24.51 \\
Autoregressive Resampling & 90.46 & 64.25 & 25.26 \\
\bottomrule
\end{tabular}
}
\end{table}

\vspace{0.5em}
\noindent\textbf{Error Simulation Strategies.}
In~\cref{sec:resampling}, we hypothesized that exposing the model to imperfect historical contexts during training mitigates error accumulation, and propose autoregressive self-resampling to simulate model errors. We compare against two alternatives: noise augmentation~\cite{valevski2024diffusion} and parallel resampling. In the first, small Gaussian noise is added to the history frames to improve robustness to inference errors. In the second, all historical frames are resampled in parallel rather than autoregressively. As shown in \cref{tab:ablation}, the autoregressive resampling strategy achieves the highest quality, followed by parallel resampling and noise augmentation. We attribute this to a mismatch between additive noise and the model’s inference-time error mode, as well as the fact that parallel resampling captures only per-frame degradation while neglecting autoregressive accumulation across time.

\vspace{0.5em}
\noindent\textbf{Simulation Timestep Shifting.} We ablate the shifting factor $s$ that biases the $t_s$ distribution. As defined in \cref{eq:timestep_shift}, a small $s$ concentrates $t_s$ in the low-noise region, and a large $s$ shifts $t_s$ toward higher noise. Equivalently, a small $s$ corresponds to weaker history resampling, encouraging faithfulness to past content, whereas a large $s$ enforces stronger resampling, promoting content modifications that may induce drift. We observe that model performance is robust to the choice of $s$; therefore, we adopt extreme values in this ablation to better visualize the impact of the shifting factor. In \cref{fig:ablation_shift}, the model trained with small $s$ exhibits error accumulation and quality degradation, while a very large $s$ reduces semantic consistency with history, increasing the risk of initial content drift. Thus, a moderate value for $s$ is essential to strike a balance between mitigating error accumulation and preventing drift.

\begin{figure}[t]
\centering \includegraphics[width=\linewidth]{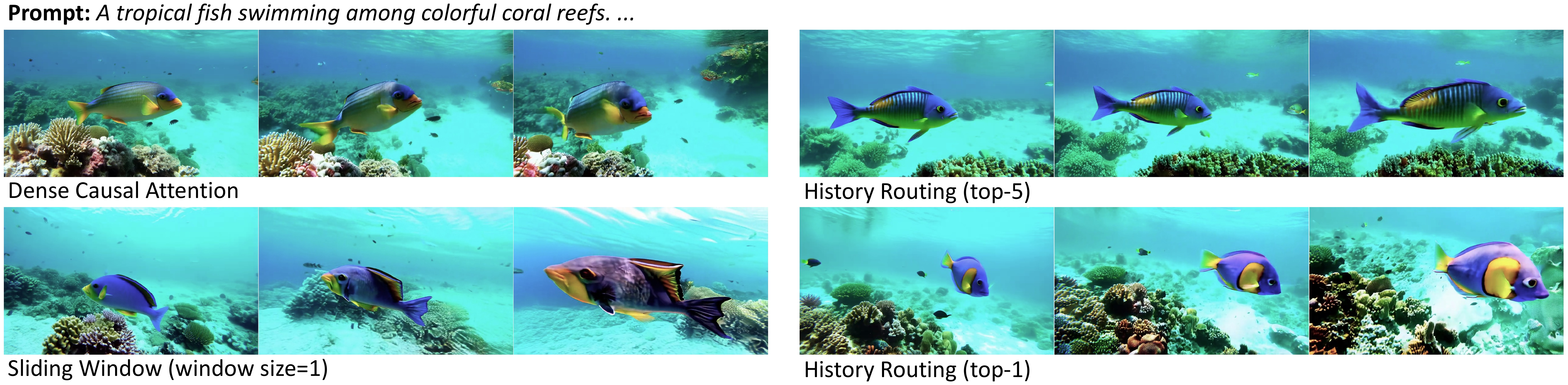}
\caption{\textbf{Sparse History Strategies.}
We compare dense causal attention, dynamic history routing, and sliding window attention in terms of appearance consistency.}
\label{fig:attention}
\end{figure}

\vspace{0.5em}
\noindent\textbf{Sparse History Strategies.} We compare three mechanisms for leveraging historical context in autoregressive generation: dense causal attention, dynamic history routing, and sliding-window attention~\cite{huang2025self, lin2025autoregressive}. We show qualitative results for dense attention, top-$5$ and top-$1$ routing, and sliding-window attention in \cref{fig:attention}. As shown in the figure, routing to the top-$5$ of $20$ history frames yields $75\%$ sparsity with quality comparable to dense attention. Reducing from top-$5$ to top-$1$ ($95\%$ sparsity) causes only minor quality degradation, demonstrating the robustness of the routing mechanism. We further contrast top-$1$ routing with a sliding window of size 1. Despite equal sparsity, the routing mechanism maintains superior consistency in the fish's appearance. We hypothesize that sliding-window attention's fixed and localized receptive field exacerbates the risk of drift. By contrast, our dynamic routing enables each query token to select diverse historical context combinations, collectively yielding a larger effective receptive field that better preserves global consistency.

\begin{figure}[t]
\centering
\includegraphics[width=0.9\linewidth]{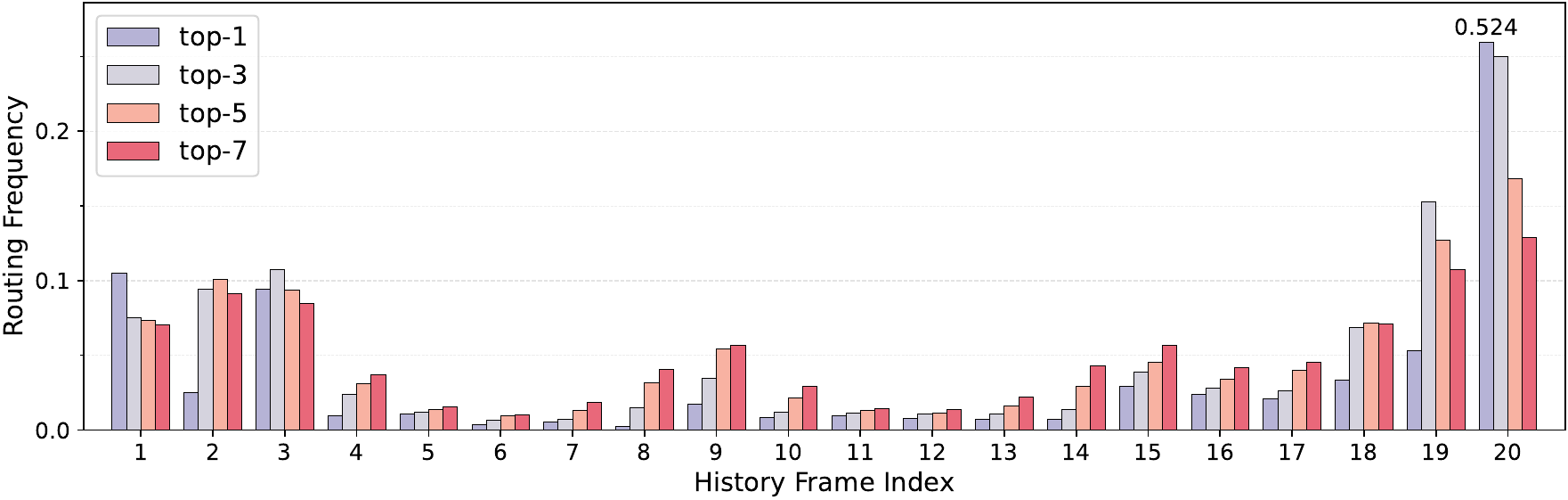}
\caption{\textbf{History Routing Frequency.}
We visualize the beginning 20 frames' frequency of being selected when generating the 21st frame. For readability, the maximum bar is truncated and labeled with its exact value. }
\label{fig:routing}
\end{figure}

\vspace{0.5em}
\noindent\textbf{History Routing Frequency.} To provide deeper insights into history routing, we experiment with $k = 1,3,5,7$ and visualize the selection frequency of each history frame during the generation of the current frame. As shown in \cref{fig:routing}, the selection frequencies exhibit a hybrid ``sliding-window'' and ``attention-sink'' pattern: the router prioritizes initial frames alongside the most recent frames preceding the target. This effect is most pronounced under extreme sparsity ($k=1$) and becomes more distributed as sparsity decreases ($k=1\rightarrow7$), encompassing a broader range of intermediate frames. The observation offers empirical support for recent attention designs combining ``frame sinks'' with sliding windows for long video rollout~\cite{yang2025longlive}, which can be viewed as a special case of our approach. Our results suggest an alternative path to attention sparsity: replacing fixed heuristic masks with a dynamic, content-aware routing mechanism capable of exploring a vastly larger space of context combinations.

\section{Discussions}

We presented \method, an end-to-end, teacher-free framework for training autoregressive video diffusion models. Identifying the root cause of error accumulation, we proposed a history self-resampling strategy that effectively mitigates this issue, ensuring stable long-horizon generation. Furthermore, we introduced a history routing mechanism designed to maintain near-constant attention complexity despite the ever-growing historical context. Experiments demonstrate our method's superior visual quality and robustness under high attention sparsity.

\clearpage
\bibliographystyle{splncs04}
\bibliography{main}
\end{document}